%% file: ral_main.tex
\documentclass[letterpaper, 10 pt, journal,twoside]{IEEEtran}
\IEEEoverridecommandlockouts

\input{configs}

\title{PR2: A Physics- and Photo-realistic Humanoid Testbed with Pilot Study in Competition}

\author{Hangxin Liu$^1$, Qi Xie$^1$, Zeyu Zhang$^1$, Tao Yuan$^1$, Song Wang$^2$, Zaijin Wang$^1$, Xiaokun Leng$^3$, \\Lining Sun$^{4,5}$, Jingwen Zhang$^1$, Zhicheng He$^3$, Yao Su$^1$

\thanks{\textit{Corresponding authors: Jingwen Zhang, Zhicheng He, and Yao Su.}}
\thanks{$^1$ State Key Laboratory of General Artificial Intelligence, Beijing Institute for General Artificial Intelligence (BIGAI), Beijing 100080, China}

\thanks{$^2$ School of Future Science and Engineering, Soochow University, Suzhou, China.}

\thanks{$^3$ Department of Computer Science, Harbin Institute of Technology, Harbin 150001, China.}

\thanks{$^4$ School of Mechatronics Engineering, Harbin Institute of Technology, Harbin 150080, China}
\thanks{$^5$ Jiangsu Provincial Key Laboratory of Advanced Robotics, School of Mechanical and Electric Engineering, Soochow University, Suzhou 215000, China.}}

\begin{document}

\maketitle
 
\begin{abstract}
This paper presents the development of a \underline{P}hysics-\underline{r}ealistic and \underline{P}hoto-\underline{r}ealistic humanoid robot testbed, PR2, to facilitate collaborative research between Embodied Artificial Intelligence (Embodied AI) and robotics. PR2 offers high-quality scene rendering and robot dynamic simulation, enabling (i) the creation of diverse scenes using various digital assets, (ii) the integration of advanced perception or foundation models, and (iii) the implementation of planning and control algorithms for dynamic humanoid robot behaviors based on environmental feedback. The beta version of PR2 has been deployed for the simulation track of a nationwide full-size humanoid robot competition for college students, attracting 137 teams and over 400 participants within four months. This competition covered traditional tasks in bipedal walking, as well as novel challenges in loco-manipulation and language-instruction-based object search, marking a first for public college robotics competitions. A retrospective analysis of the competition suggests that future events should emphasize the integration of locomotion with manipulation and perception. By making the PR2 testbed publicly available at \url{https://github.com/pr2-humanoid/PR2-Platform}, we aim to further advance education and training in humanoid robotics. 
Video demonstration: \url{https://pr2-humanoid.github.io/} 
\end{abstract}


\section{Introduction}

\begin{figure*}[ht!]
    \centering
    \includegraphics[width=\linewidth]{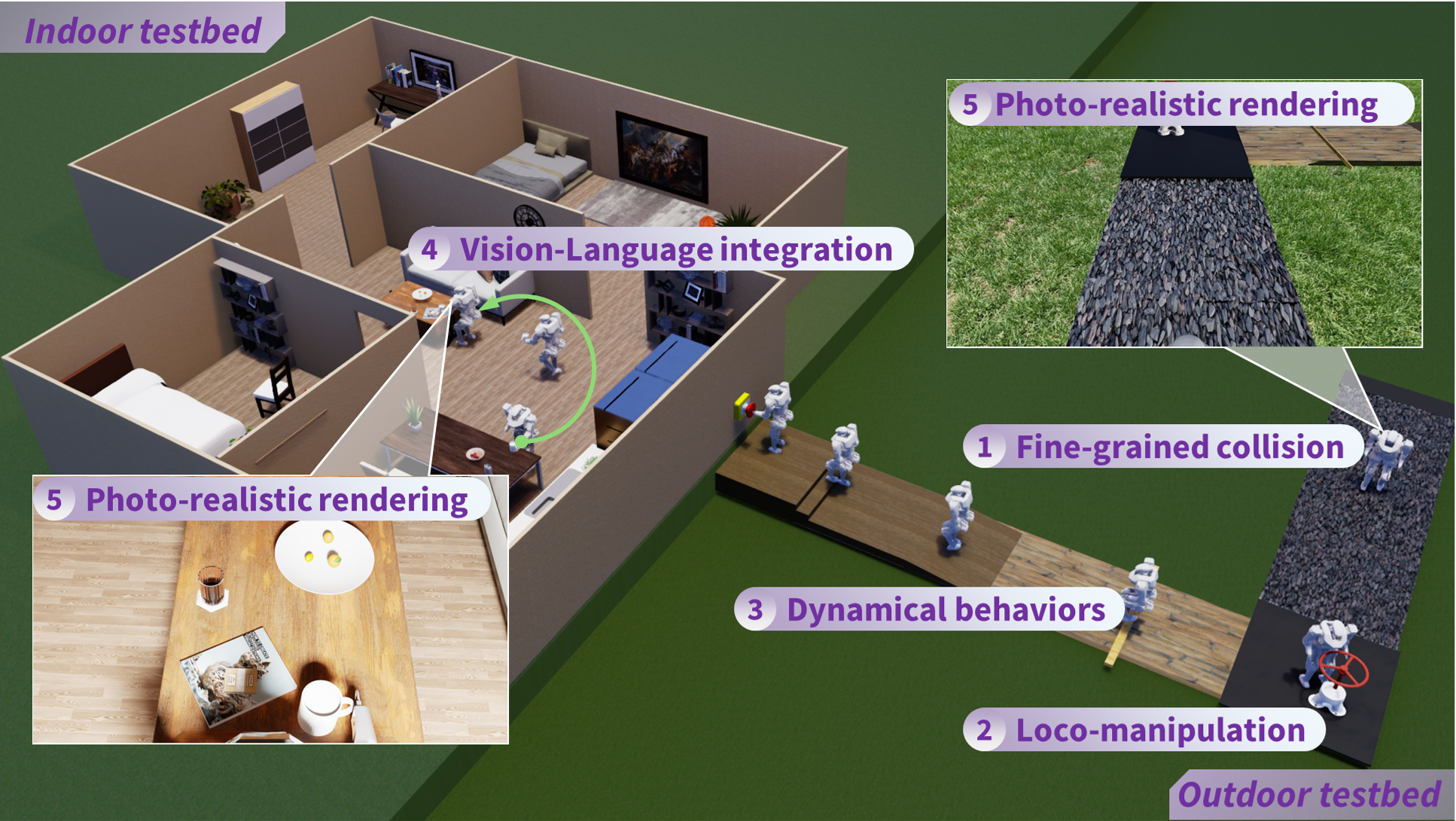}
    \caption{\textbf{The PR2 full-size humanoid robot testbed} integrates (i) photo-realistic rendering of both indoor and outdoor scenes and (ii) physics-realistic simulation of robot dynamics in locomotion and manipulation. This integration opens up new avenues for exploring the intersection of planning and control with foundation models in vision and language for humanoid robots.}
    \label{fig:teaser}
\end{figure*}

Recent academic and industrial advancements in humanoid robotics have revealed critical gaps in educational infrastructure and research accessibility. While miniature humanoid robots like DARwIn-OP~\cite{ha2011development} and NAO~\cite{gouaillier2008nao} have served as entry-level educational tools, their servo-based actuation system and limited payload capacities restrict investigations into advanced topics such as whole-body control, dynamic locomotion, and manipulation. Although emerging force/torque controlled platforms attempt to address these limitations~\cite{grimminger2020open,liu2022design}, their high maintenance requirements and substantial costs still impede widespread adoption.

Simulation environments, while not perfectly replicating the real world, have emerged as cost-effective alternatives for robotics education and research, demonstrating effectiveness across various domains~\cite{carlone2022visual,moon2024autonomous}. However, there is a significant gap in the availability of simulation platforms specifically designed for diverse topics in full-size humanoid robots. Mainstream robotics platforms including Gazebo~\cite{koenig2004design}, Webots~\cite{michel2004cyberbotics}, V-REP~\cite{rohmer2013v}, MuJoCo~\cite{todorov2012mujoco}, and PyBullet~\cite{coumans2021pybullet}, prioritize high-fidelity dynamics simulation at the expense of visual realism, thereby hindering integrated perception-action studies. Conversely, simulation environments derived from gaming engines, such as VRGym~\cite{xie2019vrgym}, ManiSkill~\cite{mu2021maniskill}, OmniGibson~\cite{li2023behavior}, Habitat~\cite{puig2023habitat}, and RoboCasa~\cite{nasiriany2024robocasa}, achieve photorealistic rendering but oversimplify physical interactions through kinematic approximations. For instance, they often assume that the robot can precisely achieve the desired pose and that the environment does not interact with the robot in unexpected ways, overlooking the fundamental challenge in humanoid robots--dynamic whole-body motions. This fundamental dichotomy creates a critical void in platforms capable of simultaneously supporting dynamic locomotion studies and vision-based manipulation tasks, as illustrated in \cref{tab:compare}.

The development landscape further compounds these challenges, as evidenced by high-profile competitions like the DARPA Robotics Challenge (DRC)~\cite{johnson2017team} and RoboCup. While these initiatives drive technological innovation, their reliance on custom-built hardware solutions and requirement for specialized expertise effectively exclude undergraduate participation and limit knowledge transfer. As a result, participants in these competitions have primarily been postgraduate students, professional engineers, and researchers.

To bridge these complementary gaps in simulation capabilities and educational accessibility, we present PR2, a novel humanoid robot testbed integrating three critical innovations: (1) Co-simulation architecture combining rigid-body dynamics with real-time photo-realistic scene rendering; (2) Modular control framework supporting hierarchical whole-body control and sensorimotor integration; (3) Multi-modal interface for foundation model integration. The efficacy of the platform was validated through a national collegiate competition involving 137 teams (402 participants) executing six progressively challenging tasks that span bipedal locomotion, multimodal manipulation, and interaction with vision and language foundation models, as shown in \cref{fig:teaser}. The four-month competition provided insight into the challenges students faced, deepened our understanding of humanoid robot education, and inspired future improvements in the affordability and accessibility of humanoid robot research and education. Following a detailed report on the main results and findings from the competition, we enhanced our testbed and made it publicly available. 
\begin{table*}[ht!]    
\centering
\caption{\textbf{Comparison of PR2 with mainstream simulation platforms}. PR2 uniquely integrates high-fidelity physics and photorealistic rendering, enabling full-body dynamic locomotion and loco-manipulation tasks. It further supports vision-language-action integration and modular control frameworks, addressing gaps in multi-modal AI-robot interaction and educational accessibility—features largely lacking in existing platforms. The platform’s co-simulation architecture and competition-ready design distinguish it as a unified testbed for humanoid research and education. The notation is defined as, (++) feature is fully available and functional, (+) feature is available, but lacking in some regards, (\textminus) feature is either available, but lacking, or only available via workarounds, (\textminus\textminus) feature is not available or difficult to integrate.}
\label{tab:compare}
\resizebox{\linewidth}{!}{
\begin{tabular}{l|c|cccccccccc}
\toprule
\textbf{Feature} & \rotatebox{45}{\textbf{PR2}} &  \rotatebox{45}{\textbf{Gazebo}} &  \rotatebox{45}{\textbf{MuJoCo}} &  \rotatebox{45}{\textbf{Webots}} &  \rotatebox{45}{\textbf{V-REP}} & \rotatebox{45}{\textbf{PyBullet}} & \rotatebox{45}{\textbf{ManiSkill}} &  \rotatebox{45}{\textbf{Habitat}} &  \rotatebox{45}{\textbf{OmniGibson}} &  \rotatebox{45}{\textbf{RoboCasa}} &  \rotatebox{45}{\textbf{VR-Gym}} \\
\midrule 
Physics Accuracy & ++ & + & ++ & + & + & + & \textminus & \textminus & + & + & \textminus \\
\hline
Photorealism & ++ & \textminus & \textminus & \textminus & \textminus\textminus & \textminus\textminus & ++ & ++ & + & ++ & + \\
\hline
Environment Creation & ++ & \textminus\textminus & \textminus\textminus & \textminus & \textminus\textminus & \textminus & + & ++ & ++ & ++ & +\\
\hline
Model Training & + & \textminus & + & \textminus & \textminus & + & + & + & + & ++ & +\\
\hline
ROS Support & + & ++ & + & ++ & ++ & + & \textminus\textminus & \textminus & \textminus & \textminus & \textminus\\
\hline
Dynamic Locomotion & ++ & ++ & ++ & + & + & + & \textminus\textminus & \textminus & \textminus & \textminus & \textminus\textminus\\
\hline
Loco-Manipulation & + & + & ++ & \textminus & \textminus & \textminus & + & \textminus & + & + & \textminus\\
\hline
Vision-Language Integration & ++ & \textminus\textminus & \textminus & \textminus & \textminus\textminus & \textminus & ++ & + & + & + & +\\
\hline
Real-Time Co-Simulation & + & ++ & ++ & ++ & ++ & ++ & + & \textminus & + & \textminus & ++\\
\hline
Humanoid Controller Integration & + & \textminus & \textminus & \textminus & \textminus & \textminus & \textminus & \textminus & \textminus & \textminus & \textminus \\
\hline
 Competition Scene \& Tasks & ++ & + & \textminus\textminus & \textminus\textminus & \textminus\textminus & \textminus\textminus & \textminus\textminus & \textminus & \textminus & \textminus & \textminus\textminus \\
\bottomrule  
\end{tabular}}
\end{table*} 

Key contributions of this work include:
\begin{itemize}
\item Open-source release of a unified simulation environment for humanoid robot education and research, addressing the realism-dynamics trade-off.
\item Curriculum framework for learning in humanoid robot perception, planning, control, and AI model integration.
\item Quantitative analysis of educational barriers in humanoid robotics through competition metrics.
\end{itemize}

\subsection{Related Work}

\subsubsection{Virtual Reality}
Contemporary virtual reality platforms have become foundational tools for developing robotics and AI simulators, leveraging advanced rendering engines to achieve photorealistic environments. Unreal Engine-based solutions such as CARLA~\cite{dosovitskiy2017carla} for autonomous cars, AirSim~\cite{shah2018airsim} for aerial vehicles, VRGym~\cite{xie2019vrgym}, and VRKitchen~\cite{gao2019vrkitchen} for robot learning demonstrate this trend, while Unity-powered platforms like AI2-THOR~\cite{kolve2017ai2}, VirtualHome~\cite{puig2018virtualhome}, and ThreeDWorld~\cite{gan2020threedworld} showcase comparable capabilities. These systems prioritize visual fidelity over physical verisimilitude, employing kinematic approximations for robot-object interactions. This architectural limitation becomes particularly problematic when simulating compliant contacts or inertial effects – critical factors in humanoid locomotion attempts to integrate basic physics engines. These findings underscore a fundamental divergence in design priorities: gaming-derived environments optimize for visual immersion, whereas robotic systems require precise dynamics resolution.

\subsubsection{Dynamic Interaction in Simulation}
The robotics community has witnessed growing adoption of NVIDIA Isaac Sim as a versatile platform for locomotion and manipulation studies, despite its utilization of conventional physics engines (\pmb{e.g.} Bullet or PhysX) shared with VR systems. This trend is exemplified by OmniGibson~\cite{li2023behavior} – the latest evolution of the Gibson environment~\cite{xia2018gibson,xia2020interactive}, and emerging \ac{eai} benchmarks including RoboCasa~\cite{nasiriany2024robocasa}, ARNOLD~\cite{gong2023arnold}, and ORBIT~\cite{mittal2023orbit}. While these platforms facilitate basic physics-based interaction modeling, they typically focus on kinematics rather than dynamics—a critical aspect for full-size humanoids. The proposed PR2 testbed enables full dynamic simulation for aggressive motions and manipulation with external objects, a key direction for future research and applications in humanoid robotics. Additionally, we have incorporated essential tools and algorithms of humanoid robots for beginners to enhance the platform's accessibility.

\subsection{Overview}
We organize the remainder of the paper as follows. In \cref{sec:system}, we introduce the system architecture of the \pmb{PR2} testbed, detailing its key modules. \cref{sec:competition} presents the competition setup based on the \pmb{PR2} testbed and describes the customizations made to support a wide range of humanoid robot tasks. \cref{sec:results} highlights the capabilities of \pmb{PR2} and summarizes the competition results. Finally, in \cref{sec:conclusion}, we conclude the paper with a discussion on the future directions for the humanoid testbed.

\begin{figure*}[t!]
    \centering
    \includegraphics[width=0.93\linewidth]{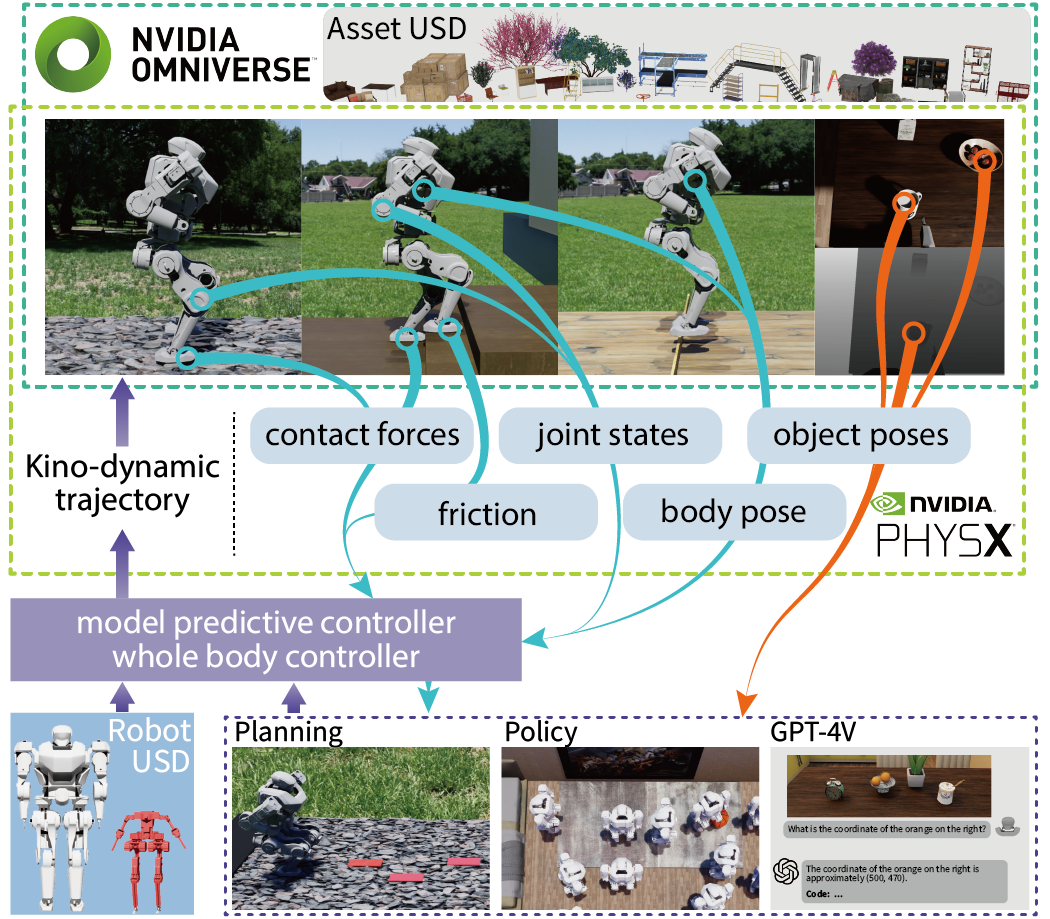}
    \caption{\textbf{System Architecture of PR2}. The testbed consists of three major modules. (i) The task module and (ii) the physics simulation module support the import of various digital assets to create diverse task environments and simulate physical effects such as forces and system dynamics. Additionally, (iii) built-in controller module includes model predictive and whole-body controllers, and high-level planners are provided for beginners but can be replaced by more advanced users as needed.}
    \label{fig:arch}
\end{figure*}

\section{System Architecture} \label{sec:system}
The system architecture of \pmb{PR2} is illustrated in \cref{fig:arch}. PR2 is built on top of NVIDIA Isaac Sim, which leverages advanced GPU-enabled graphics and physics simulation with Nvidia PhysX 5, enabling users to investigate and test robotic skills in a physics-realistic and photo-realistic simulation environment. Designed to be modular, PR2 simplifies workflows in interacting with humanoid robots with an easy-to-use interface for users who can incorporate their own modules into the platform. By extending the core components of NVIDIA Isaac Sim, PR2 consists of three main modules described in the following subsections.

\begin{figure}[ht!]
    \centering    
    \includegraphics[width=\linewidth]{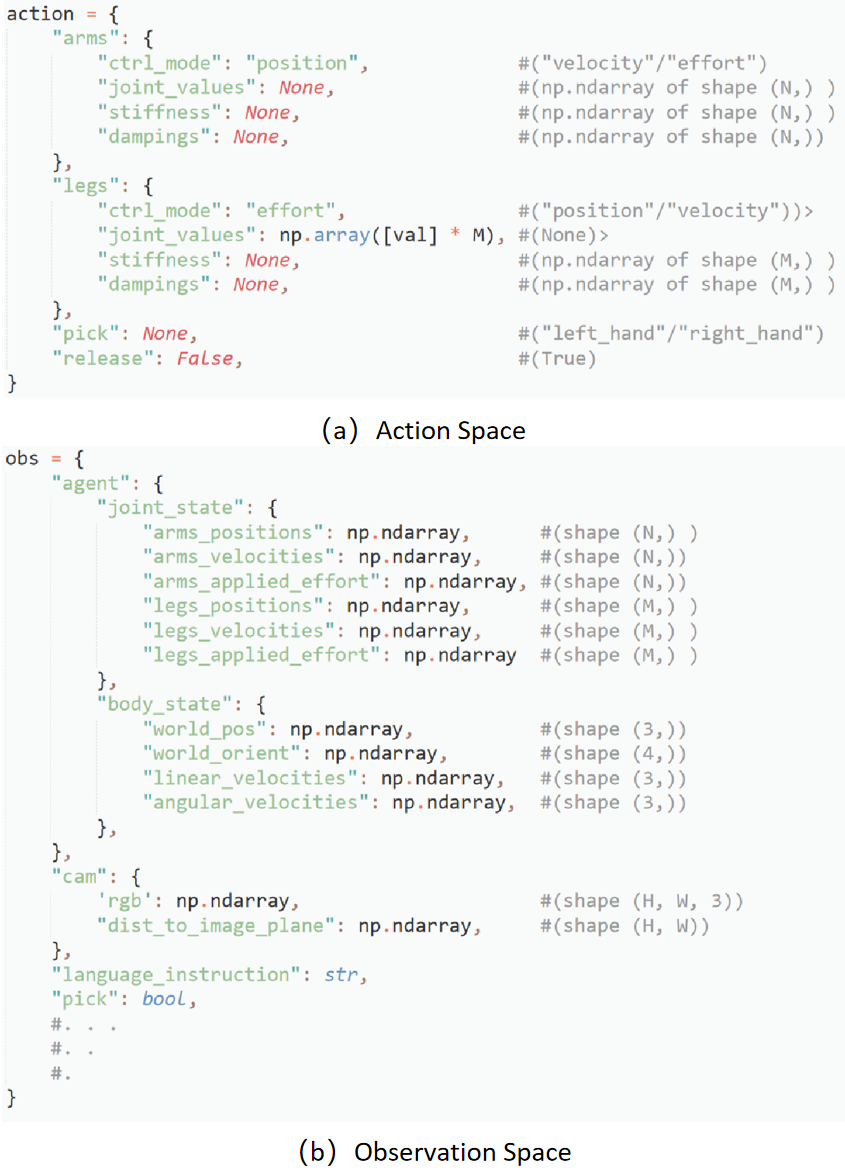}
    \caption{\textbf{The API definition for robot action and observation}. (a) The arms of the robot are position controlled while the legs are torque controlled, the manipulation actions are controlled by \texttt{pick} and \texttt{release} commands. (b) The robot's joint states and body states are provided as feedback, the \texttt{cam} topic includes the RGB image, \texttt{language\_instruction} topic supports vision-language-action tasks.}
    \label{fig:code}
\end{figure}

\subsection{Task Module}
Leveraging abundant digital assets in the form of open-source \ac{usd} files, PR2 offers a fully interactive task environment with diverse indoor and outdoor spaces populated with various objects, moving beyond the limitations of static meshes. To achieve high rendering quality, realistic ray-traced ambient light, and other lighting effects are integrated into the scene, significantly enhancing the fidelity of simulated visual sensor data. As a result, the platform can support tasks such as visual navigation, object detection, and other perception-related activities by capturing high-fidelity RGB and depth information from the robot's egocentric view. For object interactions, PR2 employs a contact sensor mechanism to determine the contact configurations between the robot and the target object.

\subsection{Physics-based Simulation}
To interface with the virtual humanoid, we adopt the API conventions of Nvidia Isaac Sim. The \texttt{Robot} class is designed to load the robot model from a \ac{usd} file and manage the associated physics handles required for setting and reading simulation states. For robot joint actuation, we offer three control modes: position, velocity, and torque. In the torque-controlled mode, users can directly apply torque commands to the robot, which are interpreted as joint efforts. In the position- and velocity-controlled modes, users specify commands for joint positions or velocities, which are then converted into joint efforts using internally implemented spring-damper models. Beyond robot actuation, we incorporate rigid body collision handling through PhysX. PR2 can also support sophisticated physical simulations, including object deformation and fluid dynamics~\cite{gong2023arnold}. This setup not only facilitates accurate robot control but also enhances the realism of interactions within the simulated environment.

\begin{figure*}[ht!]
    \centering
    \includegraphics[width=0.95\linewidth]{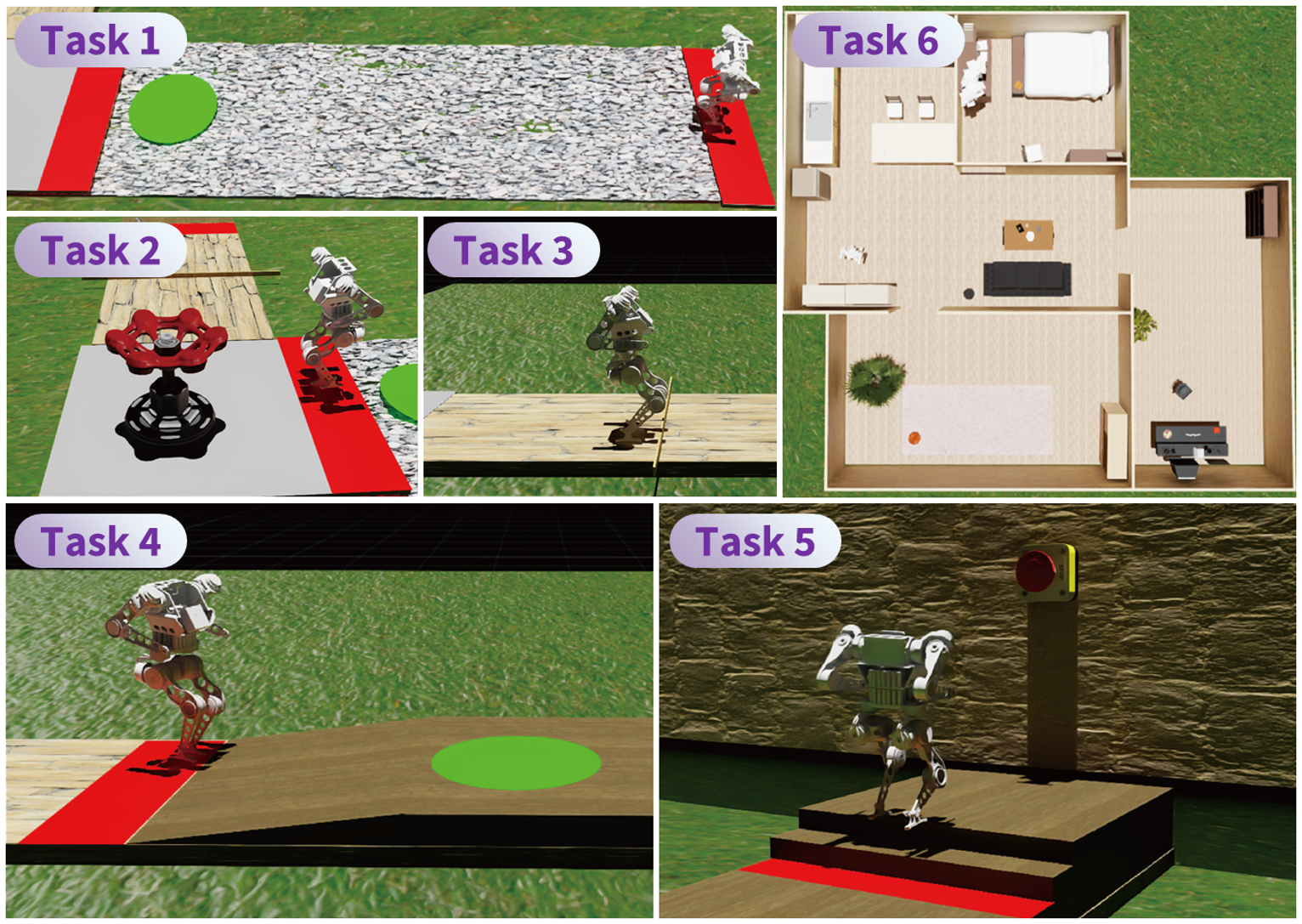}
    \caption{\textbf{Visualization of the six tasks designed for the simulation competition.} The red region indicates the initial robot placement that was randomly sampled. The tasks involving navigation are considered accomplished when the robot walks into the green region.}
    \label{fig:tasks_setup}
\end{figure*}

\subsection{Robot Controller}
Due to the underlying physics-based simulation, the humanoid imported into \pmb{PR2} would not be dynamically stable by itself. However, developing dynamic controllers of bipedal humanoids from scratch could be prohibitively difficult for beginners in this field. As a result, providing essential built-in controllers could expedite users' knowledge of such fields before implementing their own or investigating other topics on top of it.

We developed basic walking and jumping controllers in Drake~\cite{drake}, which receive user action commands, generate feedback signals from the simulator, and return the effort value of each leg joint that can be executed by the robot to the simulator through TCP/IP communication. The default parameters for these controllers would support basic stabilization and tracking but are not ideal. Users are encouraged to modify the controllers to obtain more stable and more responsive execution, and they have the option to replace these controllers with custom solutions for more sophisticated and specialized control.

An example action command is shown in \cref{fig:code}(a) where $N$ and $M$ are the number of arm and leg joints. The text in the comment indicates the type of data expected and the possible values that can be assigned to a specific field. Specifically, users can choose to actuate each robot joint by position, velocity, or effort (\pmb{i.e.}torque) based on the provided joint values array. Users can also empirically adjust the corresponding stiffness and damping for each joint, which influences the actual joint efforts applied in position- or velocity-controlled modes. For manipulation tasks, symbolic \texttt{pick} and \texttt{release} actions are designed. The \texttt{pick} action, triggered by specifying either the left or right hand, attaches the target object to the robot's end-effector if the distance between them is within a certain range. The object detaches from the robot when it receives a \texttt{True} signal for \texttt{release}.


Once an action is executed by the robot, an observation would be generated and returned to the user. An example of the observation space is provided in \cref{fig:code}(b), where $H$ and $W$ are the height and width of camera images. The observation data structure provides a comprehensive representation of the robot's state, including its joint positions, velocities, and applied efforts for both arms and legs. Additionally, the robot's body poses and its linear and angular velocities \textit{w.r.t.} the world frame are available under the \texttt{body\_state} key. The robot’s egocentric perceptual data is accessible through the \texttt{cam} key, which includes both the RGB image and the distance to the image plane (\pmb{i.e.}depth). The \texttt{language\_instruction} key enhances support for vision-language-action tasks.
A boolean value under the \texttt{pick} key indicates whether an object has been successfully attached to the robot's end-effector, a critical indicator for manipulation tasks. Additional parameters, such as stiffness and damping, or any ad-hoc specifications, can be introduced by adding new keys to the observation structure, allowing for flexible and task-specific customization.

Finally, the users are responsible for developing their high-level planners that strategize the robot's path and actions to accomplish given tasks with the required efficiency and effectiveness.

\section{Competition Overview}\label{sec:competition}

After developing the necessary modules, we organized a nationwide collegiate humanoid robot competition as a pilot study to validate the PR2 testbed through an online virtual challenge. The competition ultimately attracted 137 teams, each with up to four students. This section details the competition setup and evaluation rubric.

\subsection{Competition Trials}
To comprehensively address humanoid locomotion, manipulation, and AI integration while maintaining pedagogical accessibility, we designed six progressively challenging tasks; see \cref{tab:task_setup} for a detailed description. \cref{fig:tasks_setup} visually summarizes these tasks.

\texttt{Task~1:} As shown in the upper left of \cref{fig:tasks_setup}, the robot is initialized randomly within the red rectangular region with a heading angle deviation within $\pm30^\circ$. The objective requires the robot to traverse uneven terrain and reach the green circular target zone on the left. The task is considered successful if the robot's body moves into the circle region within $50~s$. The robot body and joint states, as well as start position and orientation, are provided as feedback, with the goal center position provided as prior information. This task tests the robustness of walking control as a basic of the humanoid robot.

\begin{table*}[ht!]
    \centering
    \caption{\textbf{Setup of the competition}. The basic walking, jumping, and step climbing controllers are provided according to task specifications. Task-related feedback is generated by the PR2 physics simulation during robot execution. \fcolorbox{black}{locomotion}{\rule{0pt}{3pt}\rule{2pt}{0pt}}, \fcolorbox{black}{manipulation}{\rule{0pt}{3pt}\rule{2pt}{0pt}}, and \fcolorbox{black}{perception}{\rule{0pt}{3pt}\rule{2pt}{0pt}} indicate score assignments related to humanoid robot's locomotion, manipulation, and perception, respectively.}
    \label{tab:task_setup}
    \resizebox{\linewidth}{!}{
        \begin{tabular}{cllll}
            \toprule
             & \textbf{Task Specification} & \textbf{Scoring Rubric} & \textbf{Feedback} &  \textbf{Require Module} \\
            \midrule 
            \multirow{3}{*}{\makecell[c]{\textbf{Task 1 (15 points)} \\ \textbf{Uneven terrain walking}}} &   \multirow{3}{*}{goal center position} & \cellcolor{locomotion}walk into the task region (+5) & \multirow{3}{*}{\makecell[l]{robot body \& joint states\\start position \& orientation}}  & \multirow{3}{*}{walking velocity planner} \\
            & & \cellcolor{locomotion}walk half long (+5) & & \\
            & & \cellcolor{locomotion}walk into finish region (+5) & & \\
            \midrule
            \multirow{3}{*}{\makecell[c]{\textbf{Task 2 (15 points)} \\ \textbf{Valve turning}}} &   \multirow{3}{*}{\makecell[l]{valve center position\\valve dimensions\\valve ID}} & \cellcolor{locomotion}walk into the task region (+5) & \multirow{3}{*}{\makecell[l]{robot body \& joint states\\start position \& orientation\\arm pick state}}  & \multirow{3}{*}{\makecell[l]{loco-manipulation \\ planner and controller}} \\
            & & \cellcolor{manipulation}turn the valve (+5) & & \\
            & & \cellcolor{manipulation}turn the valve more than $45^{\circ}$ (+5) & & \\
            \midrule 
            \multirow{3}{*}{\makecell[c]{\textbf{Task 3 (15 points)} \\ \textbf{Vertical jumping}}} & \multirow{3}{*}{moving bar ID} & \cellcolor{locomotion}step over the moving bar (+10) & \multirow{3}{*}{\makecell[l]{robot body \& joint states\\start position \& orientation\\ moving bar position \& velocity}}  &  \multirow{3}{*}{jumping motion planner} \\
            & & \cellcolor{locomotion}jump over the moving bar (+15) & & \\
            & & contact with the moving bar (-5) & & \\
            \midrule 
            \multirow{2}{*}{\makecell[c]{\textbf{Task 4 (10 points)} \\ \textbf{Slope climbing}}}& \multirow{2}{*}{\makecell[l]{goal center position\\slope incline angle}} & \cellcolor{locomotion}walk into the task region (+5) &  \multirow{2}{*}{\makecell[l]{robot body \& joint states\\start position \& orientation}}  &  \multirow{2}{*}{\makecell[l]{slope walking controller}}  \\
            & & \cellcolor{locomotion}walk into the finish region (+5) & & \\
            \midrule
            \multirow{4}{*}{\makecell[c]{\textbf{Task 5 (20 points)} \\ \textbf{Step climbing} \\ \textbf{and button pressing}}} & \multirow{4}{*}{\makecell[l]{button center position\\step dimensions}} & \cellcolor{locomotion}walk onto the step (+5) &  \multirow{4}{*}{\makecell[l]{robot body \& joint states\\start position \& orientation}}  &  \multirow{4}{*}{\makecell[l]{step walking planner\\walking velocity planner\\arm motion planner}} \\
            & & \cellcolor{locomotion}walk over all the steps (+5) & & \\
            & & \cellcolor{manipulation}touch the button (+5) & & \\
            & & \cellcolor{manipulation}touch the button with arm (+5) & & \\
            \midrule
            \multirow{4}{*}{\makecell[c]{\textbf{Task 6 (25 points)} \\ \textbf{Indoor object search} \\ \textbf{and transport}}}  & \multirow{4}{*}{\makecell[l]{3D model of 9 objects\\ coffee table center position \\ a sentence of task:\\ put the object on the coffee table}} & \cellcolor{perception}walk to the pick region (+5) &  \multirow{4}{*}{\makecell[l]{robot body \& joint states\\ start position \& orientation\\arm pick state\\camera RGB-D information}} &  \multirow{4}{*}{\makecell[l]{loco-manipulation controller \\  Vision-Language-Action model}} \\
            & & \cellBG{double color fill={manipulation}{perception}, shading angle=-45}{l}{pick up target object (+10)} & & \\ & & \cellcolor{locomotion}walk to the table (+5) & & \\
            & & \cellcolor{manipulation}place the object on the table (+5) & & \\
            \bottomrule  
        \end{tabular}}%
\end{table*}

\texttt{Task~2:} Illustrated in the middle left of \cref{fig:tasks_setup}, the robot needs to turn the red valve to more than $45^\circ$ within $4~mins$ to be considered successful, which is a loco-manipulation task. For simplification of contact dynamics (details provided in \cref{sec:testbed}), a fixed joint is automatically established between the robot's end-effector and the valve when their proximity falls below a threshold, by sending the \texttt{pick} command.

\texttt{Task~3:} The center plot of \cref{fig:tasks_setup} illustrates the task specifications, where a bar moves horizontally from right to left, requiring a jumping movement of the robot over the bar and landing stably within $4~mins$ to acquire full points. The position and velocity of the bar are provided as feedback for jumping motion planning and control. This task tests the capability of humanoid robots to conduct highly dynamic motions.

\texttt{Task~4:} The lower left of \cref{fig:tasks_setup} demonstrates the initialization area and task goal region of this task. The robot needs to walk through a slope within $2~mins$, with the provided slope incline angle as prior.

\texttt{Task~5:} The robot needs to first walk through both steps and the plain floor and then contact the red button within $4~mins$ to accomplish this task (see lower right plots of \cref{fig:tasks_setup}). The dimensions and number of steps are provided as the task prior.  

\texttt{Task~6:} This is a comprehensive indoor object search task (lower right of \cref{fig:tasks_setup}), where a sentence of the task will be provided at first to describe the task. Three possible objects are designed: the cup on the kitchen table shown on the upper left, the bear on the table in the playroom (mid bottom), and the cube on the desk of the studyroom (right bottom), these objects are randomly initialized in the rectangular region of each room. The robot needs to find the assigned object and translate it to the table in the living room within $4~mins$. The 3d model of each object is provided as prior, and the camera view of the robot is streamed out as feedback.    

\begin{figure*}[t!]
    \centering   
    \includegraphics[width=0.85\linewidth]{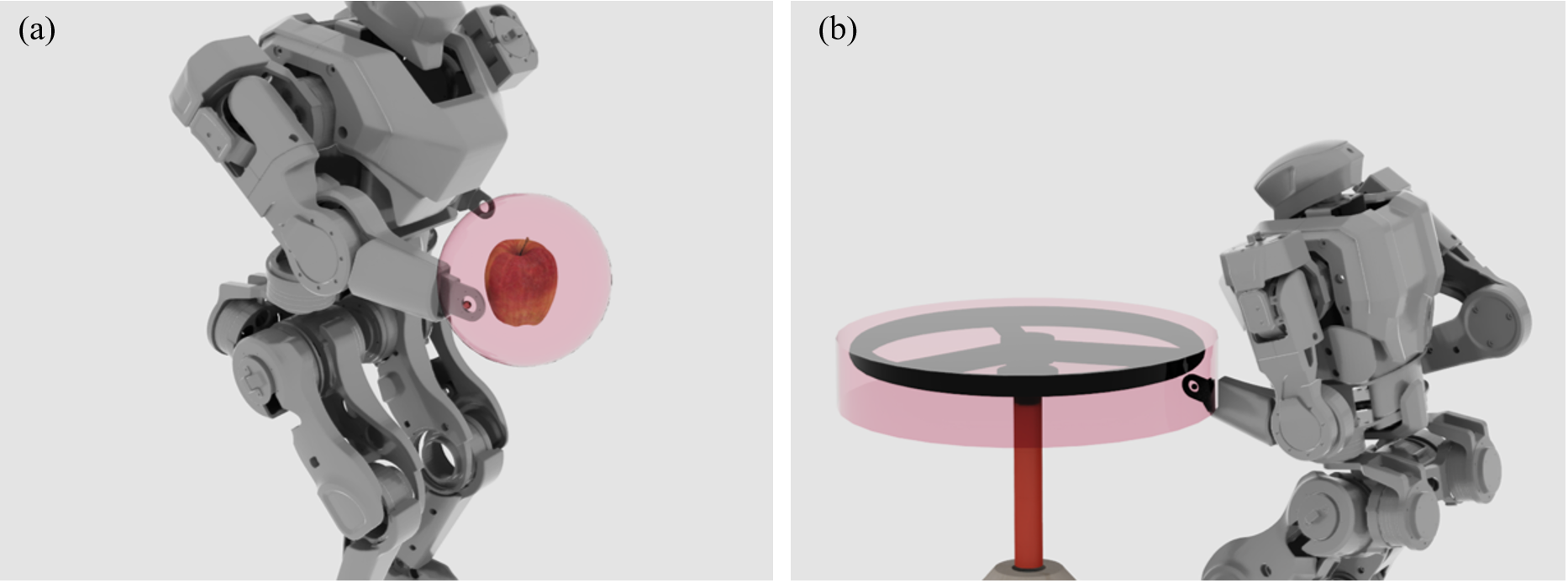}
    \caption{\textbf{Contact establishment for manipulation tasks}. In loco-manipulation tasks, the robot can enter object manipulation mode when an intersection is detected between the volume of the robot arm's distal end and the enlarged object meshes, as indicated by the light red color. (a) The articulated valve is represented as an enlarged short cylinder, while (b) the object to be grasped is modeled as a sphere.} 
    \label{fig:grasping}
\end{figure*}

These tasks are operated independently rather than sequentially, which ensure that participants who may struggle with one aspect still have opportunities to engage with other challenges. For example, difficulties in manipulation in \texttt{Task 2} do not impact the performance in subsequent tasks involving locomotion, such as jumping (\texttt{Task 3}) and climbing slopes and stairs (\texttt{Task 4} and \texttt{5}). For each task, if the robot falls, walks outside the task region, or reaches the assigned time limit, the simulation will terminate automatically. 

\begin{table}[t!]
    \centering
    \caption{\textbf{Main Physical Parameters of KUAVO v1.0 Robot}}
    \label{tab:hardware}
    \resizebox{0.95\linewidth}{!}{%
        \begin{tabular}{c  c  c  c  c}
            \toprule
            \multicolumn{5}{c}{\textbf{Dimension Parameters}} \\
            \midrule
            Total mass & Total height & Pelvis width & Thigh length & Calf length \\
            34.5 [kg] & 1.20 [m] & 0.22 [m] & 0.23 [m] & 0.26 [m]  \\
            Foot length & Upper arm length & Forearm length \\
            0.15 [m] & 0.24 [m] & 0.20 [m] \\
            \toprule
            \multicolumn{5}{c}{\textbf{Motion Range \& Joint Peak Torque}}\\
            \midrule
            Hip Yaw & Hip Roll & Hip Pitch & Knee Pitch &  Ankle Pitch\\
            $-90^{\circ} \sim 60^{\circ}$ & $-30^{\circ} \sim 75^{\circ}$ &  $-30^{\circ} \sim 120^{\circ}$ & $-120^{\circ} \sim 10^{\circ}$ & $-30^{\circ} \sim 80^{\circ}$\\
            48 [Nm] &110 [Nm] &110 [Nm] &110 [Nm] &48 [Nm] \\
            \bottomrule
        \end{tabular}}%
\end{table}

\subsection{The Robot}
The KUAVO v1.0 platform, a humanoid robot with a height of $1.2~m$ and a weight of $34.5~kg$~\cite{he2024cdm}, is used in the virtual competition. It incorporates 18 \acp{dof}, 5 for each leg and 4 for each arm. In this competition, all joint actuators of the KUAVO robot operate in the torque-controlled mode, better replicating the control architecture of a full-size humanoid. For the details of the hardware platform's specifications, please refer to \cref{tab:hardware}.

\begin{figure*}[t!]
    \centering    
    \includegraphics[width=0.8\linewidth]{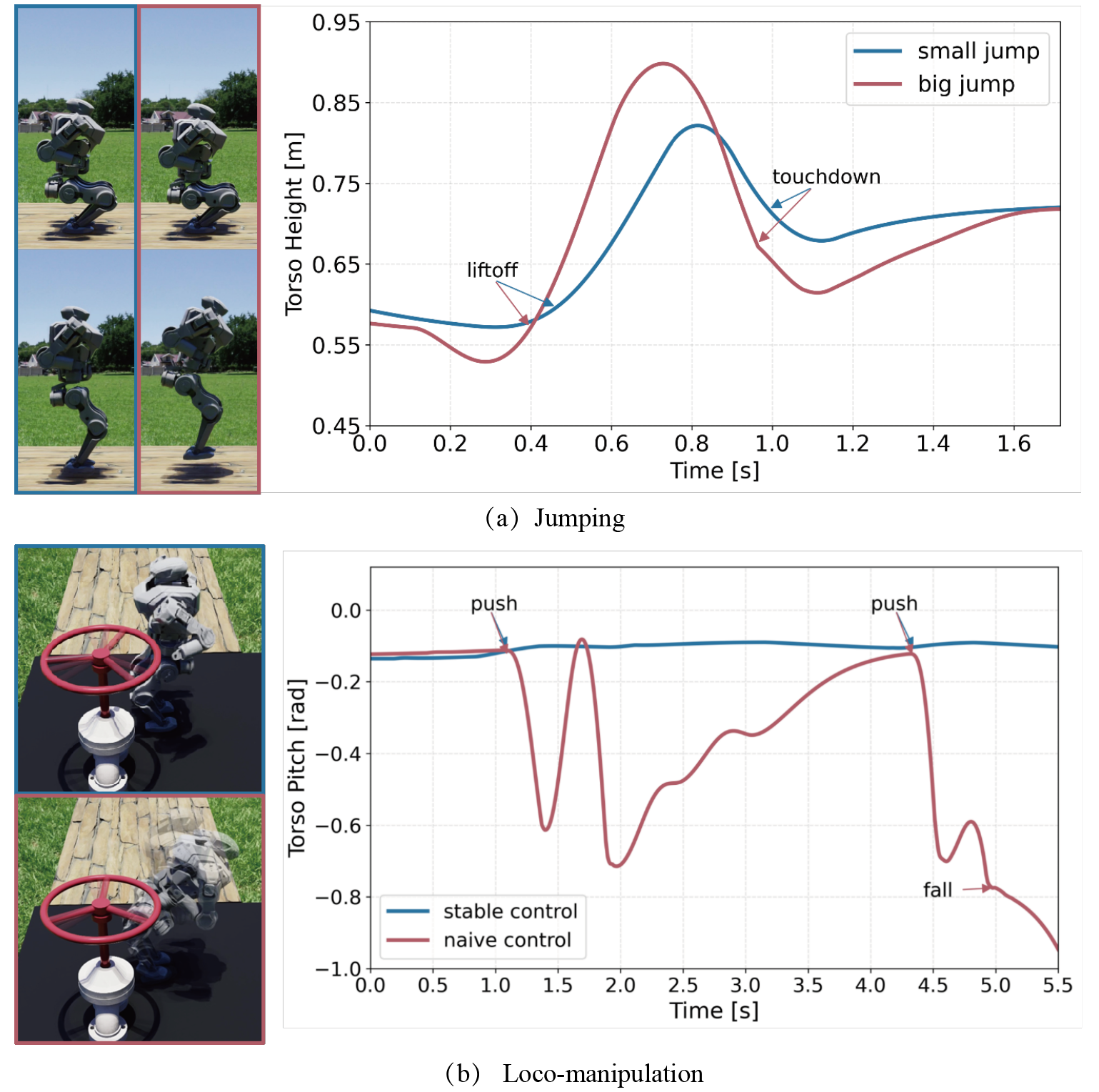}
    \caption{\textbf{The physics-realistic simulation on humanoid's dynamical behaviors}. (a) The robot jumps to a significantly different height by a slight difference in crouching, reflecting the importance of supporting dynamical behaviors in the humanoid testbed. (b) The reactive force introduces a strong disturbance to the robot when it pushes the valve. PR2 can produce different pushing outcomes depending on the robustness of the controller implemented in the robot. }
    \label{fig:dynamic_capability}
\end{figure*}
\subsection{The Virtual Testbed} \label{sec:testbed}
The PR2 utilizes the \texttt{Articulations} package from the Isaac Sim Core extension to achieve realistic dynamic behaviors through whole-body torque control on the KUAVO robot. The \texttt{Articulations} package provides a high-level interface for managing robots using the Root Articulation API, facilitating effective handling of their attributes and properties.

To ensure stability during simple standing as well as complex tasks such as traversing rough terrains or performing loco-manipulation, three basic controllers are provided for the KUAVO during the competition: (i) a standard walking controller that accepts body velocity commands (3-\acp{dof}) for Tasks 1, 2, 4, and 6; (ii) a dynamic jumping controller based on a centroidal dynamics model~\cite{he2024cdm}, which takes inputs for jump time, velocity, and height for Task 3; and (iii) a step walking controller based on the Zero Moment Point (ZMP) method~\cite{scianca2020mpc} for Task 5. The arm controller is not provided and must be developed by the teams to complete Tasks 2, 5, and 6 successfully. Critical implementation details for each task are outlined below:

\begin{itemize}[leftmargin=*,noitemsep,nolistsep,topsep=0pt]
    \item \textbf{\texttt{Task~1}}: The surface of the $4.96$-meter-long rocky road features fine-grained collision geometry that results in unstable footsteps with different contact forces applied to the robot.
    \item \textbf{\texttt{Task~2}}: The users could turn the valve (diameter around $0.54~m$) either by hitting it with the robot arm's distal end or by establishing a fixed attachment with the valve when the distal end reaches the enlarged valve volume (see \cref{fig:grasping}(b)) by sending a ``pick'' command; this allows the valve revolves together with the robot's locomotion.
    \item \textbf{\texttt{Task~3}}: Users can either design a gait that steps over the bar and adjust the built-in walking controller to track the gait stably or use the provided jumping motion planner to jump over it.
    \item \textbf{\texttt{Task~4}}: This task is similar to \texttt{Task~1}, but the users need to tune the planner and controller to climb the slope with a $7^{\circ}$ incline.

    \item \textbf{\texttt{Task~5}}: Users must design a gait for the robot to ascend stairs ($0.1~m$ high and $2~m$ wide) while maintaining the desired heading direction to press a button. The pressing mechanism is implemented using the PhysX Contact Report API to detect if the robot (preferably its arm) makes contact with the button's surface.
    \item \textbf{\texttt{Task~6}}: The robot would need visual perception (to recognize scenes and objects), language understanding (to translate questions and instructions into actions), and navigation in complex environments (to move and find things). When the user sends a pick command, a fixed joint will be created only if the Euclidean distance between the end-effector and the object to be picked is less than $0.2~m$, expressed as a spherical region as shown in \cref{fig:grasping}(a).
\end{itemize}

For each task, the robot's initial pose is randomly sampled within a designated start region. Although the objective interactions in task 2 and task 6 have been simplified for the college competition, the underlying PhysX engine is fully capable of supporting more advanced manipulation tasks. For example, complex interactions, such as grasping objects with a multi‐finger gripper, can be achieved by utilizing finer‐grained collision-checking and more sophisticated control strategies. It is important to note that while these capabilities exist within our simulation framework, the explicit demonstration remains one of our ongoing efforts in future PR2 releases.

\begin{figure*}[ht!]
    \centering
    \includegraphics[width=0.95\linewidth]{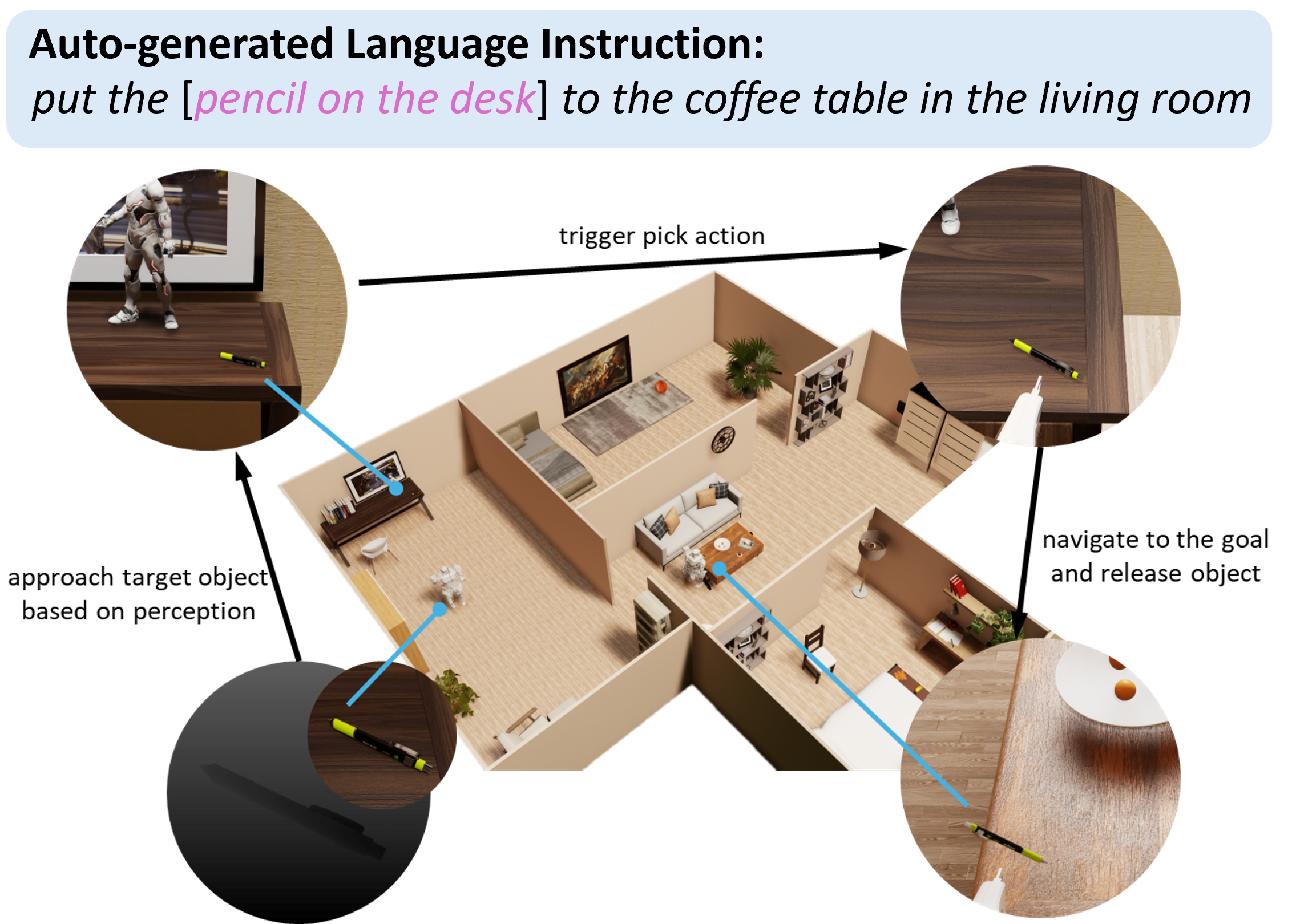}
    \caption{\textbf{Vision and language-based object rearrangement}. Based on the parsed input language instruction, the robot must first detect the object based on its image/depth observations and navigate toward it before picking and transporting it.}
    \label{fig:vla}
\end{figure*}

\section{Results}\label{sec:results}
The beta version of PR2 was released to participants in January 2024, with the program submission deadline set for May 20, 2024. The competition ultimately received 51 valid submissions. In this section, we first highlight three unique features of PR2. Then, we report and analyze the competition results. Finally, we summarize our key takeaway from the competition with critical lessons learned.

\subsection{Platform Capability}

\subsubsection{Kinodynamic Locomotion}
Leveraging the underlying PhysX engine and the Isaac Sim APIs, PR2 enables the execution of generated kinodynamic motion plans. With built-in controllers, \cref{fig:dynamic_capability}(a) shows the robot performing vertical jumping to a moderate height and to an extreme height that is pushing the controller to its limit. Such jumping trajectories are naturally kinodynamic due to the launching,  flight, and landing phases during the jump all pose unique challenges in dynamics planning and control, and all require high-fidelity dynamical simulation to capture and reflect such processes. 

Specifically, (i) after generating trajectories for the center of mass (CoM) and centroidal angular momentum (CAM) within the hardware capabilities, PR2 needs to accurately compute the robot's dynamical state evolvement. (ii) In the flight phase, the platform must determine the controller's trajectory tracking performance to reflect the disturbances from execution errors. (iii) As the robot lands from a higher jump, it experiences larger impact forces. Thus, PR2 must not only reflect different landing impacts but also propagate such impacts to the landing controller to determine robot stability. \cref{fig:dynamic_capability}(a) demonstrates the execution of two significantly different jump trajectories, wherein the red one takes longer for the controller to stabilize the robot after landing. The capability to reflect the subtle differences in locomotion is a critical aspect of a humanoid testbed.

\subsubsection{Physics-based Interaction}
Together with the realistic simulation of robot locomotion, PR2 further enables humanoid's physics-based interaction with the environment during manipulation. Similar to the jumping example, \cref{fig:dynamic_capability}(b) shows how the reactive force caused by the target valve influences robot motions in loco-manipulation. With a naive controller, the reactive force causes a dramatic torso pitch, whereas a more stable controller could successfully reject the reactive force and maintain body balance.

\subsubsection{VLA Interface}
We introduce customized APIs that enable users to interact with large language models (LLMs) and vision-language models (VLMs), providing a new approach to exploring AI and robotics. In \texttt{Task 6}, users receive visual observations and language instructions to identify goals, formulate plans, and compute robot actions (\pmb{e.g.} joint positions or efforts) at each timestep. The target object in the language instruction is randomly selected from a set in PR2, and the robot is initialized at a random location in the same room as the target object.

\cref{fig:vla} illustrates a typical example where the instruction is \textit{put the [pencil on the desk] on the coffee table in the living room}. The task involves both object detection (identifying the \textit{pencil}) and scene understanding (locating the \textit{coffee table}). Typically, the pencil is in a different room from the coffee table, requiring the user to generate high-level navigation and motion plans to guide the robot in avoiding obstacles and executing the necessary actions. The agent must take the \textit{pick} action when it believes it is close enough to the target object. If the distance is within a specified threshold, a fixed joint is created to attach the target object to the robot’s end effector. Upon reaching the coffee table, the agent must take the \textit{release} action to complete the task.

\subsection{Team Performance Analysis}

\begin{figure*}[ht!]
    \centering
    \includegraphics[width=0.75\linewidth,trim=0.5cm 0.8cm 0cm 0cm,clip]{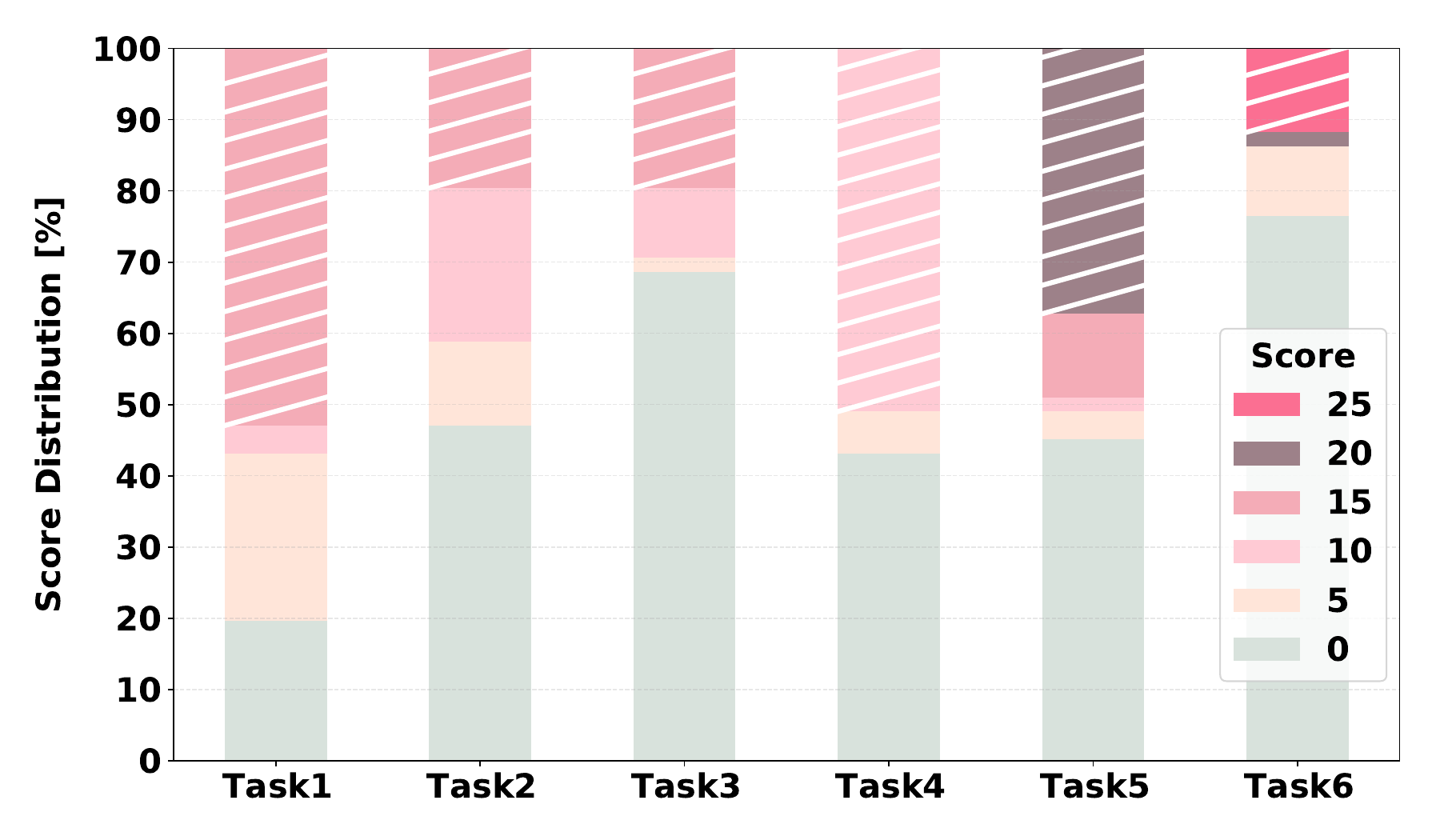}
    \caption{\textbf{Performance statistics}. Each segment represents the percentage of participants whose scores align with the rubric in \cref{tab:task_setup}. Segments with white stripes indicate the number of participants who achieved full marks for the task.}
    \label{fig:score_result}
\end{figure*}
\begin{figure*}[t!]
    \centering
    \includegraphics[width=0.9\linewidth]{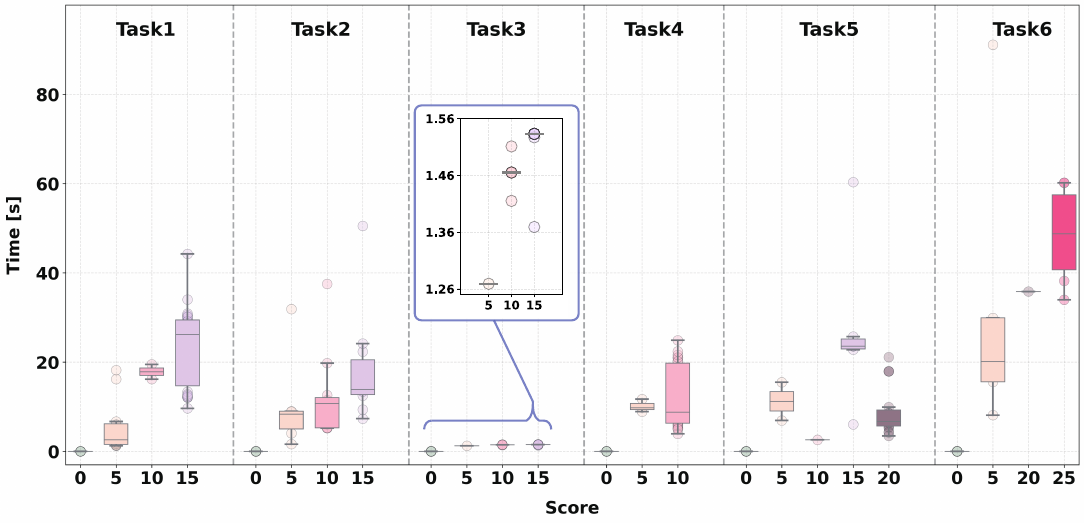}
    \caption{\textbf{Box plots of completion times across six tasks, categorized by scores received}. The distribution of completion times varies significantly across tasks, with notable outliers in several tasks, highlighting the diversity in participant performance and the varying difficulty levels of the tasks.}
    \label{fig:time_result}
\end{figure*}

\begin{figure*}[t!]
    \centering
    \includegraphics[width=0.75\linewidth]{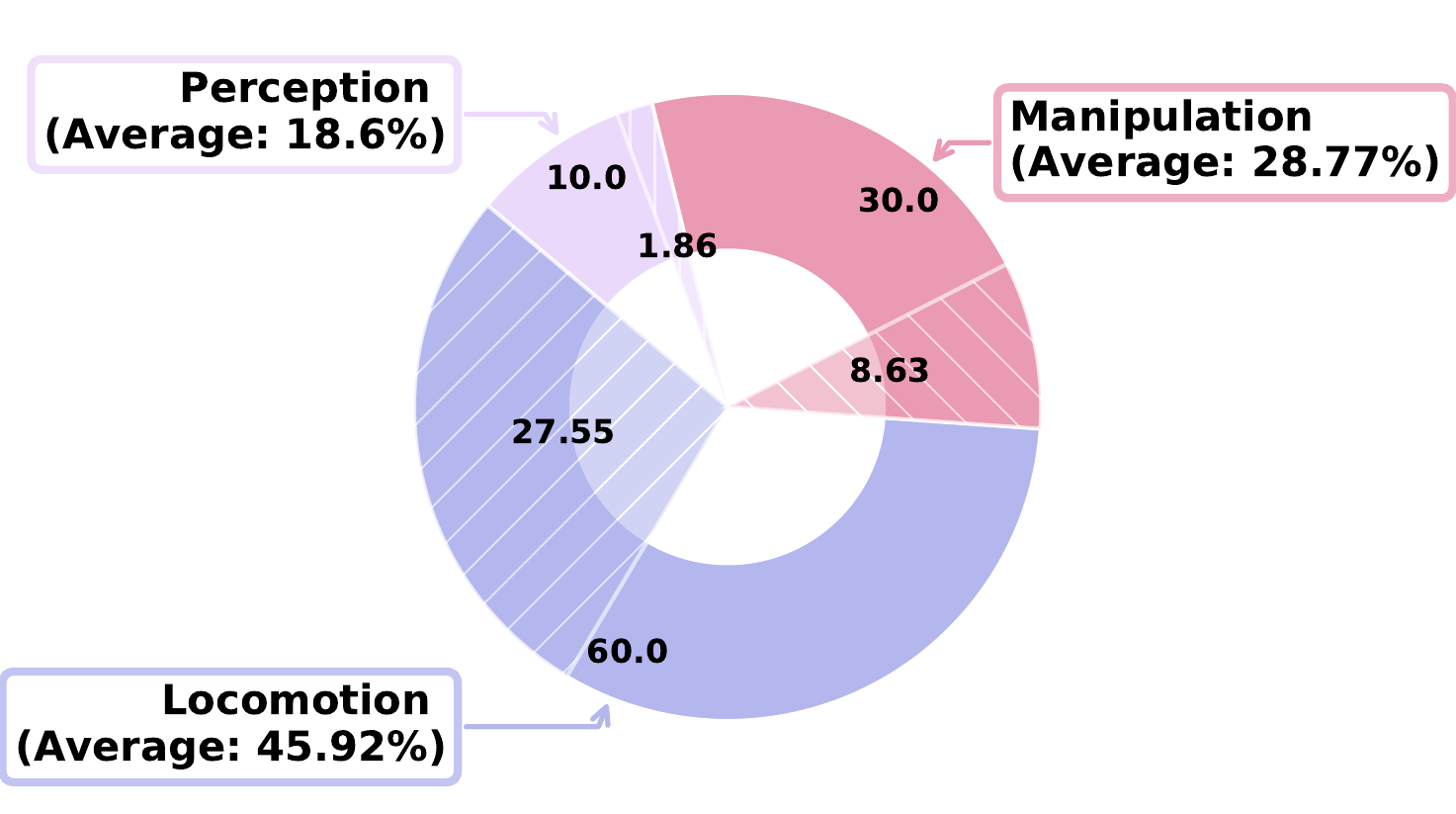}
    \caption{\textbf{Average performance in three categories of Tasks}. The solid slices in the chart represent the total available points for each category, while the dashed areas indicate the average points scored by all participants.} 
    \label{fig:score_pie_chart}
\end{figure*}

\cref{fig:score_result} presents the score distribution percentages across the six tasks. Each bar representing a task is divided into colored segments corresponding to different score ranges. 

For \texttt{Task 1}, the majority of participants (52.94\%) achieved the full score of 15, followed by 23.53\% scoring 5, and a small portion (3.92\%) received a score of 10. \texttt{Task 2} exhibited a more varied distribution due to the increased difficulty of balancing bipedal walking with the constraints of the contacted valve. Approximately 58.82\% of participants scored between 0 and 5 (indicating an inability to turn the valve), while only 19.61\% fully succeeded in this task. In \texttt{Task 3}, despite the provided jumping controller, nearly 70\% of participants failed to score any points. As the only task involving a dynamically changing environment (\pmb{i.e.}the moving bar), participants struggled to design a high-level plan for the robot to jump and land at the right moments without hitting the bar. 
\texttt{Task 4} and \texttt{5}, both of which emphasized bipedal walking capabilities for climbing slopes and stairs, respectively, showed similar success rates to Task 1: 50.98\% of participants in \texttt{Task 4} and 37.25\% in \texttt{Task 5} achieved full marks.
\texttt{Task 6}, which integrated perception, language parsing, manipulation, and walking, presented the greatest variation in scores, with only 11.76\% of participants successfully completing it. The performance diversity across tasks highlighted the importance of introducing different challenges in competition.

\cref{fig:time_result} further presents the individual time consumption for each score across all tasks, with each score represented by a distinct box plot that shows the distribution of times achieved by participants. The overall plot reveals varying levels of difficulty and completion times for each task. Generally, the robot requires more time as it becomes more involved in the task to achieve higher scores. Task 6 exhibits the widest range of scores and the most balanced distribution, indicating diverse levels of student performance. This task saw some students excelling, while others struggled, as evidenced by the highest proportion of participants scoring 0 and the second highest scoring 25. The presence of outliers across all tasks suggests that while most students completed within a typical range, some took significantly longer.

\begin{figure*}[t!]
    \centering
    \includegraphics[width=\linewidth]{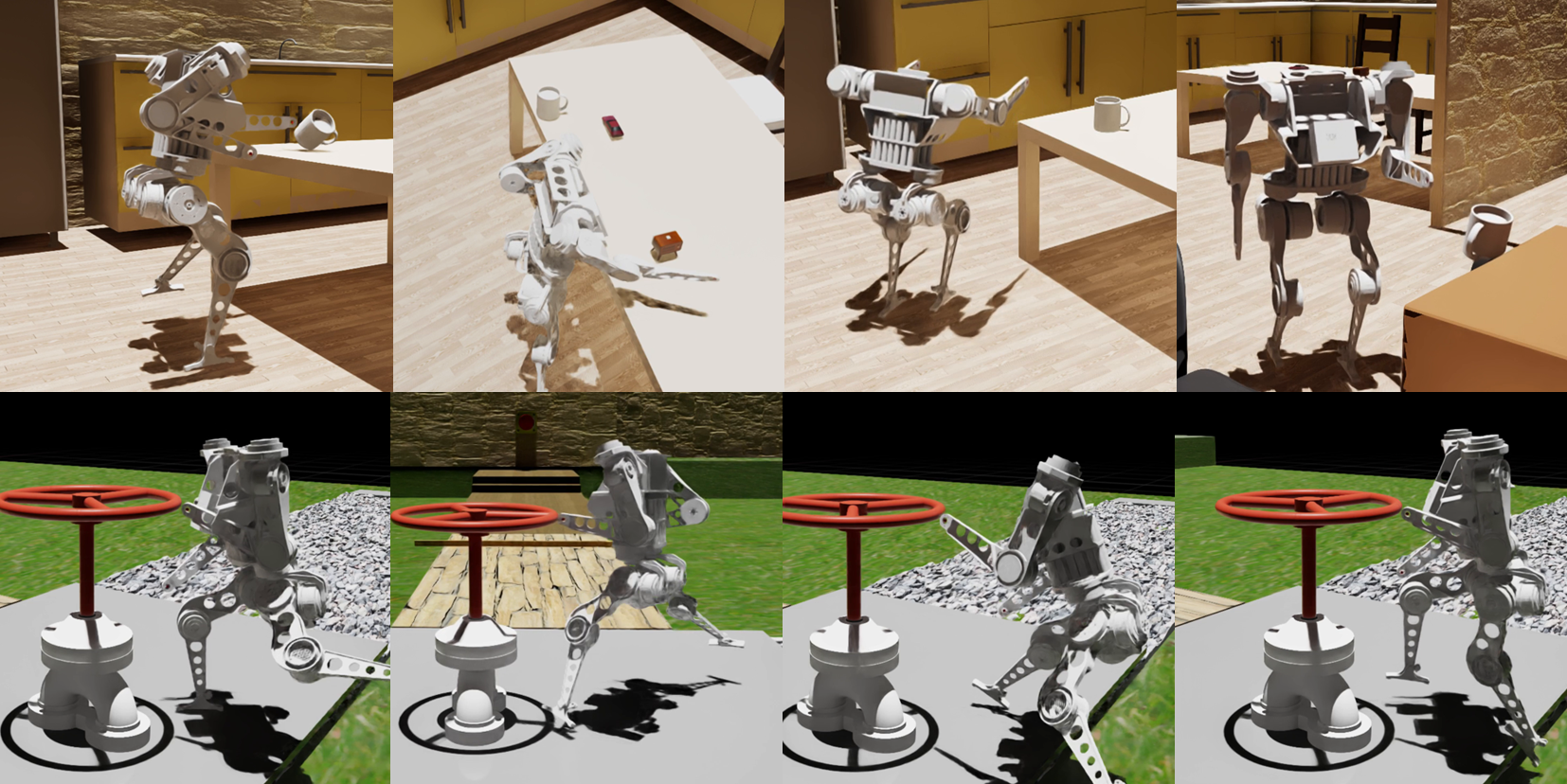}
    \caption{\textbf{Main reasons for failure}. (Top row) Prior to manipulating the target object, the robot must accurately detect its pose through its perception system. Incorrect detection and collisions with the environment were the primary causes of failure. (Bottom row) The robot struggled to maintain stability due to inadequate balancing between lower-body locomotion and upper-body manipulation while turning the valve.}
    \label{fig:failure}
\end{figure*}

By categorizing the detailed score rubric into three categories—locomotion (60\%), manipulation (30\%), and perception (10\%)—as shown in \cref{tab:task_setup}, we analyzed the average points participants scored (see \cref{fig:score_pie_chart}). The results indicate that participants performed relatively well in locomotion, earning nearly half of the available points (27.55 out of 60), reflecting the emphasis this area has received in humanoid literature. However, they encountered significantly greater challenges in tasks involving manipulation and perception, scoring less than 30\% and 20\% of the available points in these categories, respectively. \cref{fig:failure} shows some typical failure cases in manipulating the valve (\texttt{Task 2}) and perception (\texttt{Task 6}).

\begin{figure*}[t!]
    \centering
    \includegraphics[width=\linewidth]{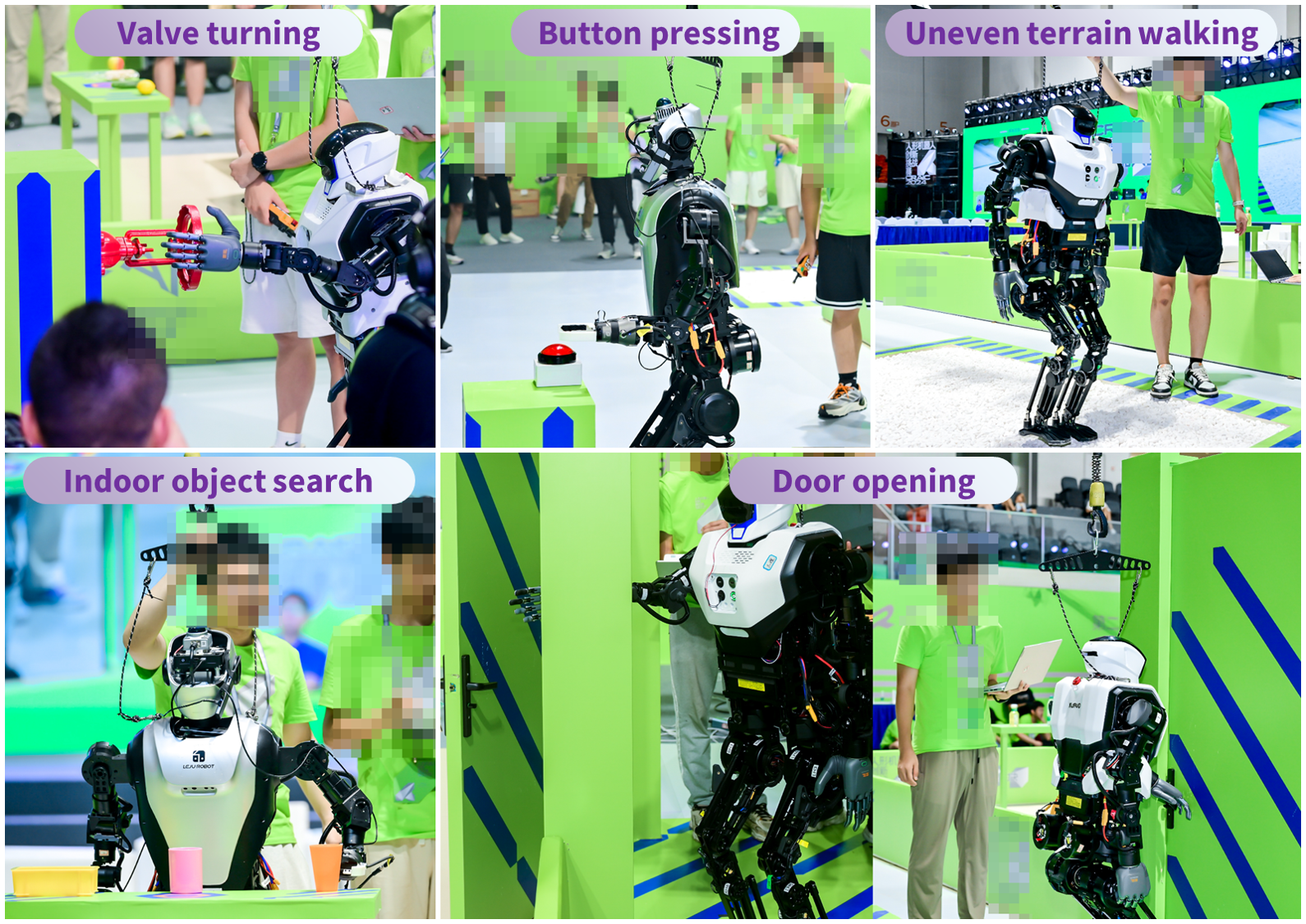}
    \caption{\textbf{Images of the actual physical humanoid robot competition.} The participants were allowed to develop and test their strategies in PR2 prior to the physical competition to reduce time consumption in deployments.}
    \label{fig:physical}
\end{figure*}

\subsection{Lesson Learned}
Throughout the competition, we hosted an online forum to address participant questions and technical issues. Combined with the competition results, this process provided valuable insights that guided our engineering efforts and may also benefit the broader robotics community in advancing research and education on humanoid robots.

\subsubsection{Trilemma of rendering quality, simulation precision, and computing resource}

Despite the rapid advancements in AI and robotics, access to dedicated Nvidia GPUs remains limited, particularly for students from traditional engineering backgrounds. During the competition, PR2 encountered over 20 issues related to platform launch failures, primarily due to missing or inadequate GPUs. Unfortunately, some participants were forced to withdraw due to the lack of GPU access.

In the competition, a limited number of participants had access to high-end consumer-grade GPUs, such as the RTX 40 Series, which can cost up to $\$2000$. To address these hardware constraints, we introduced new APIs that enable participants to adjust rendering quality and modify the simulation time-step. Reducing rendering quality can increase the frame rate, ensuring that PR2 can photo-realistically reflect the simulated physics. We conducted benchmarks on a computer equipped with a 13th Gen Intel® Core™ i9-13900KF CPU and an Nvidia 4090 GPU. With a fixed physical simulation time-step of 0.001$s$, \cref{tab:fps-vs-res} indicated that the frame rate remains stable until the rendering workload approaches the system's computational limits. Beyond a resolution of $1280\times720$ pixels, the frame rate decreases as rendering demands exceed available resources, necessitating adjustments to maintain consistent physics time-steps. Alternatively, increasing the simulation time-step can alleviate computational burdens. Larger time-steps give the physics engine coarser time slices to compute interactions, including the integration of forces and motion, leading to numerical errors and degraded fidelity. Balancing rendering quality, simulation precision, and computing resources is a crucial design consideration for promoting more accessible and equitable education opportunities in humanoid robotics.

\begin{table}[!t]
\centering
\caption{\textbf{Average viewpoint frames per second (FPS) at different resolutions}. The average FPS is calculated by dividing the total number of frames (30,161) by the total duration of the recording.}
\label{tab:fps-vs-res}
\resizebox{0.95\linewidth}{!}{
\begin{tabular}{lcccc}
\toprule
 \textbf{Resolution (pixels)} & \textbf{640$\times$480} & \textbf{1024$\times$768} & \textbf{1280$\times$720} & \textbf{1600$\times$1200}  \\
 \midrule
\textbf{Average FPS} & 17.363 & 17.683 & 17.361 & 16.263\\
\midrule\midrule
\textbf{Resolution (pixels)} & \textbf{1920$\times$1080} & \textbf{2560$\times$1440} & \textbf{3440$\times$1440} & \textbf{3840$\times$2160} \\
\midrule
\textbf{Average FPS} & 14.918 & 15.033 & 13.973 & 13.697 \\
\bottomrule
\end{tabular}}
\end{table}

\subsubsection{Model-based controller v.s. Model-free policy}
Although the competition provided basic controllers for robot stabilization and tracking, some participants opted to develop and train RL-based policies instead. While both approaches can effectively generate robot motions and are encouraged, they introduce significant challenges in (i) creating a robust testbed that supports both model-based and model-free methods, potentially combining elements of them in hybrid approaches, and (ii) ensuring fair evaluation as RL often require significant computational resources for training.

\subsubsection{The necessity and challenges in integrating AI-related topics}
Participants generally performed well in locomotion-related tasks with the provided controllers, as shown in \cref{fig:score_pie_chart}. However, significant performance drops were observed when tasks incorporated elements of manipulation and visual or language perception, as reflected by lower average scores. Specifically, difficulties were most pronounced in \texttt{Task 3} and \texttt{6}, which required high-level planning and the integration of AI techniques. These results underscore the need for future research and education in humanoid robotics to encompass a broader range of skills and topics, addressing the challenges that arise when integrating complex AI components with robot tasks.

\subsubsection{Competition with Physical Robots}\label{sec:real}
The top 10 teams from the simulation competition were invited to participate in the physical humanoid robot competition held in August 2024. Five trials were designed based on the simulated tasks in PR2, with modifications to accommodate physical robots (see \cref{fig:physical}). For the valve-turning and button-pressing tasks, the robot needed to locate the target object, approach it, and use its arm to either rotate a horizontal valve 180 degrees or press a button. In the uneven terrain walking task, the robot had to maintain stability while navigating rough surfaces. The door-opening task required the robot to grasp and rotate a doorknob. In the indoor object search task, an audio cue indicated the target object the robot needed to fetch, incorporating challenges in speech recognition, object detection, locomotion, and grasping. Each team had one hour for competition and participants could complete the tasks in any order, but bonus points were awarded for completing tasks consecutively.

Several sim-to-real gaps were identified during the competition. Accurately replicating the damping and friction characteristics of both the robot's actuators and the valves used in the physical competition within the simulation environment proved challenging. This discrepancy can lead to differences in robot performance between simulated and real-world scenarios. Additionally, the simulation environment often fails to capture the complex lighting variations present in real-world settings, affecting the robot's vision system and its ability to accurately perceive and interact with objects. Despite the substantial domain gap between simulation and real-world execution, PR2 remains an effective testbed for participants to become familiar with the robotic platform and build a foundation for the competition. On average, participants scored $75\%$, $80\%$, $63\%$, and $83\%$ in the valve-turning, button-pressing, uneven terrain walking, and door-opening tasks, respectively—significantly higher than the average performance in simulation competition. However, the indoor object search task saw an average score of only $5\%$, highlighting the challenges of integrating locomotion with manipulation and visual/audio perception.

\section{Conclusion and Looking Forward}\label{sec:conclusion}
In this paper, we presented the development of PR2, a testbed that combines physics-realistic simulation with photo-realistic scene rendering. By integrating high-quality scene rendering with high-fidelity simulation of robot dynamics, PR2 offers a new platform for exploring full-size humanoid robot locomotion, manipulation, and perception, addressing critical aspects for future humanoid applications. The deployment of PR2 in an online collegiate competition attracted hundreds of participants, revealing the challenges junior students faced in combining manipulation and perception in humanoid robot locomotion and emphasizing the need for future research and education to focus on these complex integrations. We have released an upgraded version of the PR2 testbed, which we believe will significantly benefit future research and education in humanoid robotics, particularly for those who previously lacked access. Moving forward, our priorities include incorporating a built-in RL training scheme alongside planning and control methods, enhancing scene diversity through scene synthesis techniques~\cite{yang2024physcene}, and refining fine-grained grasping capabilities for loco-manipulation tasks in future PR2 releases.

\paragraph*{Acknowledgement:}
This work was supported in part by the National Natural Science Foundation of China (No. 62403064, 62403063). We thank the engineering team from Leju Robot for technical support, Ms. Zhen Chen from BIGAI for refining the figures, and all colleagues from the BIGAI TongVerse project for fruitful discussions and help with simulation developments.

{
\bibliographystyle{ieeetr}
\bibliography{reference}
}

\end{document}

%% file: configs.tex
\usepackage{anyfontsize}
\usepackage{graphicx} \graphicspath{ {figures/} }
\usepackage{amsmath,amssymb,mathabx}
\usepackage{algorithmic}
\usepackage[ruled,linesnumbered]{algorithm2e}
\usepackage{acronym}
\usepackage{enumitem}
\usepackage{booktabs}
\usepackage{hyperref}
\usepackage{bm}
\usepackage{balance}
\usepackage{xspace,setspace}
\usepackage[skip=3pt,font=small]{subcaption}
\usepackage[skip=3pt,font=small]{caption}
\usepackage[capitalise]{cleveref}
\usepackage{tabularx,colortbl,multirow,array,makecell}
\usepackage{overpic}
\usepackage{cite}
\usepackage{tikz}
\usepackage[]{mdframed}
\usepackage{xcolor,verbatimbox}
\catcode`>=\active %
\catcode`<=\active %

\catcode`>=12 %
\catcode`<=12 %

\makeatletter
\DeclareRobustCommand\onedot{\futurelet\@let@token\@onedot}
\def\@onedot{\ifx\@let@token.\else.\null\fi\xspace}

\makeatother

\makeatletter
\def\BState{\State\hskip-\ALG@thistlm}
\makeatother

\makeatletter
\renewcommand{\paragraph}{%
  \@startsection{paragraph}{4}%
  {\z@}{0ex \@plus 0ex \@minus 0ex}{-1em}%
  {\hskip\parindent\normalfont\normalsize\bfseries}%
}
\makeatother

\crefname{algorithm}{Alg.}{Algs.}
\Crefname{algocf}{Algorithm}{Algorithms}
\crefname{section}{Sec.}{Secs.}
\Crefname{section}{Section}{Sections}
\crefname{table}{Tab.}{Tabs.}
\Crefname{table}{Table}{Tables}
\crefname{figure}{Fig.}{Fig.}
\Crefname{figure}{Figure}{Figure}

\definecolor{gblue}{HTML}{4285F4}
\definecolor{gred}{HTML}{DB4437}

\acrodef{rl}[RL]{Reinforcement Learning}
\acrodef{il}[IL]{Imitation Learning}
\acrodef{dof}[DoF]{Degree of Freedom}
\acrodef{ik}[IK]{Inverse Kinematics}
\acrodef{mpc}[MPC]{model predictive control}
\acrodef{wbc}[WBC]{whole-body control}
\acrodef{ros}[ROS]{Robot Operating System}
\acrodef{slam}[SLAM]{Simultaneous Localization and Mapping}
\acrodef{com}[CoM]{center of mass}
\acrodef{cam}[CAM]{centroidal angular momentum}
\acrodef{cmm}[CMM]{centroidal momentum matrix} 
\acrodef{cdm}[CDM]{centroidal dynamics model}
\acrodef{cop}[CoP]{Center of Pressure}
\acrodef{usd}[USD]{Universal Scene Description}
\acrodef{eai}[Embodied AI]{Embodied Artificial Intelligence}

\newcolumntype{x}{>{\columncolor{gblue}}l}
\newcolumntype{y}{>{\columncolor{gred}}l}
\newcolumntype{z}{>{\columncolor{LightOliveGreen}}l}

\definecolor{locomotion}{RGB}{179, 183, 238}  
\definecolor{manipulation}{RGB}{234, 154, 178} 
\definecolor{perception}{RGB}{235, 217, 252}   

\SetKwComment{Comment}{//}{}

\usepackage{tikz}
\usetikzlibrary{tikzmark, calc}
\tikzset{
    double color fill/.code 2 args={
        \pgfdeclareverticalshading[%
            tikz@axis@top,tikz@axis@middle,tikz@axis@bottom%
        ]{diagonalfill}{100bp}{%
            color(0bp)=(tikz@axis@bottom);
            color(50bp)=(tikz@axis@bottom);
            color(50bp)=(tikz@axis@middle);
            color(50bp)=(tikz@axis@top);
            color(100bp)=(tikz@axis@top)
        }
        \tikzset{shade, left color=#1, right color=#2, shading=diagonalfill}
    }
}
\newcommand\BGcell[4][0pt]{%
  \begin{tikzpicture}[overlay,remember picture]%
    \path[#4] ( $ (pic cs:#2) + (-.5\tabcolsep,1.9ex) $ ) rectangle ( $ (pic cs:#3) + (\tabcolsep,-#1*\baselineskip-.8ex) $ );
  \end{tikzpicture}%
}%
\newcounter{BGnum}
\newcommand\cellBG[3]{
    \multicolumn{1}{
        !{\BGcell{startBG\arabic{BGnum}}{endBG\arabic{BGnum}}{%
                #1}
            \tikzmark{startBG\arabic{BGnum}}}
            #2
        !{\tikzmark{endBG\arabic{BGnum}}}}
        {#3} 
      \addtocounter{BGnum}{1}
}

\tikzset{%
    diagonal fill/.style 2 args={%
        double color fill={#1}{#2},
        shading angle=45,
        opacity=0.8},
    other filling/.style={%
        shade,
        shading=myshade,
        shading angle=0,
        opacity=0.5}
   }